%% file: acl_latex.tex
\documentclass[11pt]{article}

\usepackage[final]{acl}

\usepackage{times}
\usepackage[T1]{fontenc}
\usepackage{CJKutf8}
\usepackage{amsmath}
\usepackage{amsfonts}
\usepackage{booktabs}
\usepackage{multirow}
\usepackage{microtype}
\usepackage{inconsolata}
\usepackage{tabularx}
\usepackage{graphicx}
\usepackage{subcaption}
\usepackage{xcolor}
\usepackage[most]{tcolorbox}

\definecolor{noambcolor}{HTML}{4C78A8}
\definecolor{textrescolor}{HTML}{59A14F}
\definecolor{visioncolor}{HTML}{E15759}

\newcommand{\amb}[1]{%
  \colorbox{yellow!35}{\strut #1}%
}

\newtcolorbox{examplepanel}[1]{
  enhanced,
  colframe=#1,
  colback=white,
  colbacklower=#1!5,
  boxrule=0.7pt,
  arc=1.5mm,
  left=1.5mm,
  right=1.5mm,
  top=1.5mm,
  bottom=1.5mm,
  segmentation style={solid,draw=#1!45,line width=0.5pt},
  before skip=0pt,
  after skip=0pt,
  equal height group=labelExamplesFig,
  valign=top,
  valign lower=top,
  space to lower
}

\title{When Are Images Needed in Multimodal Machine Translation? Text-Only Routing for On-Demand Visual Context}

\author{
    \textbf{Xinyu Wang\textsuperscript{1}}, 
    \textbf{Yingfeng Luo\textsuperscript{1}}, 
    \textbf{Yuxin Gao\textsuperscript{1}},
    \textbf{Anxiang Ma\textsuperscript{1,2}\thanks{Corresponding author.}}, 
    \textbf{Jingbo Zhu\textsuperscript{1,2}},
    \textbf{Tong Xiao\textsuperscript{1,2}}
    \\
    \textsuperscript{1} School of Computer Science and Engineering, Northeastern University, Shenyang, China\\
    \textsuperscript{2} NiuTrans Research, Shenyang, China\\
    \texttt{2601866@stu.neu.edu.cn}\\
	\texttt{\{maanxiang,zhujingbo,xiaotong\}@mail.neu.edu.cn}
}

\begin{document}
\maketitle

\input{pages/sections/0_abstract}

\input{pages/sections/1_introduction}

\input{pages/sections/2_relatedwork}

\input{pages/sections/3_Methodology}

\input{pages/sections/4_Dataset}

\input{pages/sections/5_Experiment}

\input{pages/sections/6_Result_and_Analysis}

\input{pages/sections/7_Conclusion}

\input{pages/sections/8_Limitations}

\bibliography{references}

\clearpage

\input{pages/sections/Appendix}

\end{document}

%% file: pages/sections/0_abstract.tex
\begin{abstract}
Image-guided multimodal machine translation (MMT) leverages visual context to resolve translation-relevant uncertainty. However, prevailing systems either process images unconditionally or apply gating only after visual encoding, leaving redundant visual computation unavoidable. To mitigate this computational overhead, routing decisions must rely on a pre-visual source signal. Crucially, surface-level ambiguity does not equate to visual dependence, as intra-sentence textual context often suffices for disambiguation. We therefore formulate visual invocation as source-text sufficiency prediction: determining exclusively from the source text whether translation-relevant uncertainty remains unresolved. To support this formulation, we construct a visual-dependency dataset with three fine-grained labels---\texttt{no\_ambiguity}, \texttt{text\_resolved}, and \texttt{vision\_needed}---and verify discriminative visual evidence for \texttt{vision\_needed} instances. We further propose PreVIR, a lightweight, backend-agnostic text-only router that selects between text-only and multimodal branches before computing visual features. Experiments on English-to-Chinese MMT demonstrate that PreVIR achieves a favorable COMET--budget trade-off relative to strong routing baselines. At its default operating point, it substantially reduces visual calls while maintaining translation quality comparable to always-on multimodal inference. Label-wise and case analyses show that this trade-off stems from distinguishing genuine visual dependency from contextually resolved ambiguity.
\end{abstract}

%% file: pages/sections/1_introduction.tex
\section{Introduction}
\begin{figure*}[t]
    \centering
    \begin{subfigure}[t]{0.315\textwidth}
        \caption{\texttt{no\_ambiguity}}
        \label{fig:no_ambiguity_example}

        \begin{examplepanel}{noambcolor}
            \centering
            \includegraphics[
                width=\linewidth
            ]{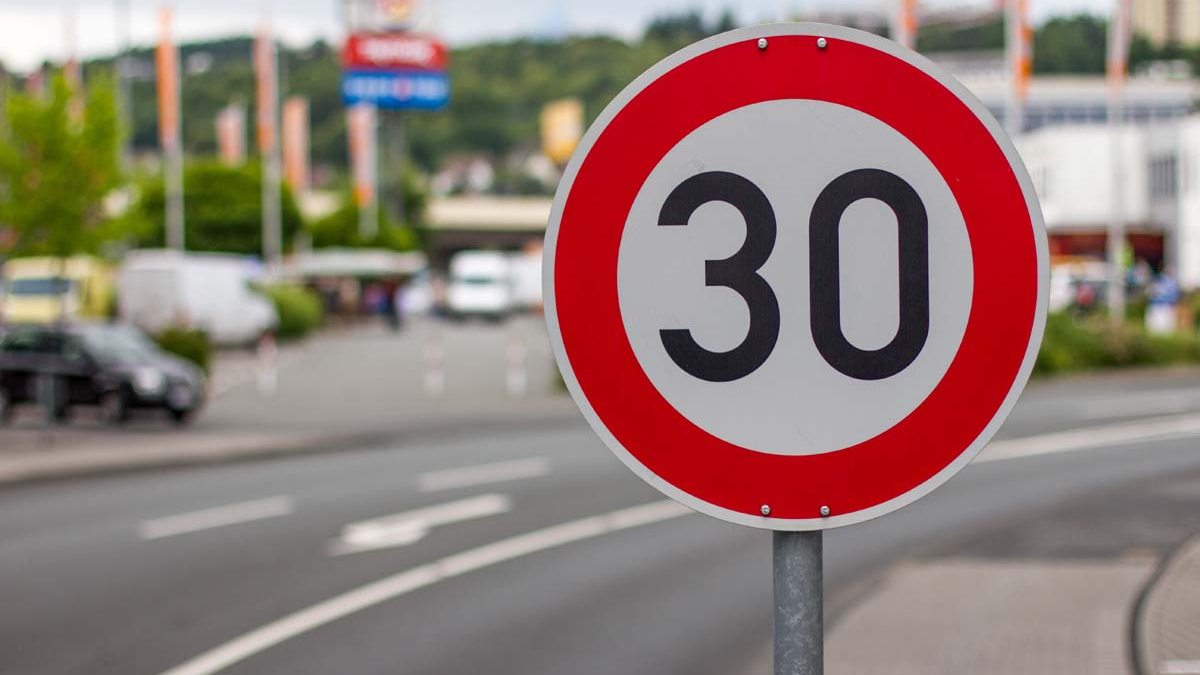}

            \tcblower

            \raggedright
            \small
            \textbf{Source:}
            The speed limit is 30.

            \vspace{1mm}
            \textbf{Ambiguity:}
            None.

            \vspace{1mm}
            \textbf{Resolving evidence:}
            The source is self-contained.

            \vspace{1mm}
            \textbf{Visual role:}
            Not required.
        \end{examplepanel}
    \end{subfigure}
    \hfill
    \begin{subfigure}[t]{0.315\textwidth}
        \caption{\texttt{text\_resolved}}
        \label{fig:text_resolved_example}

        \begin{examplepanel}{textrescolor}
            \centering
            \includegraphics[
                width=\linewidth
            ]{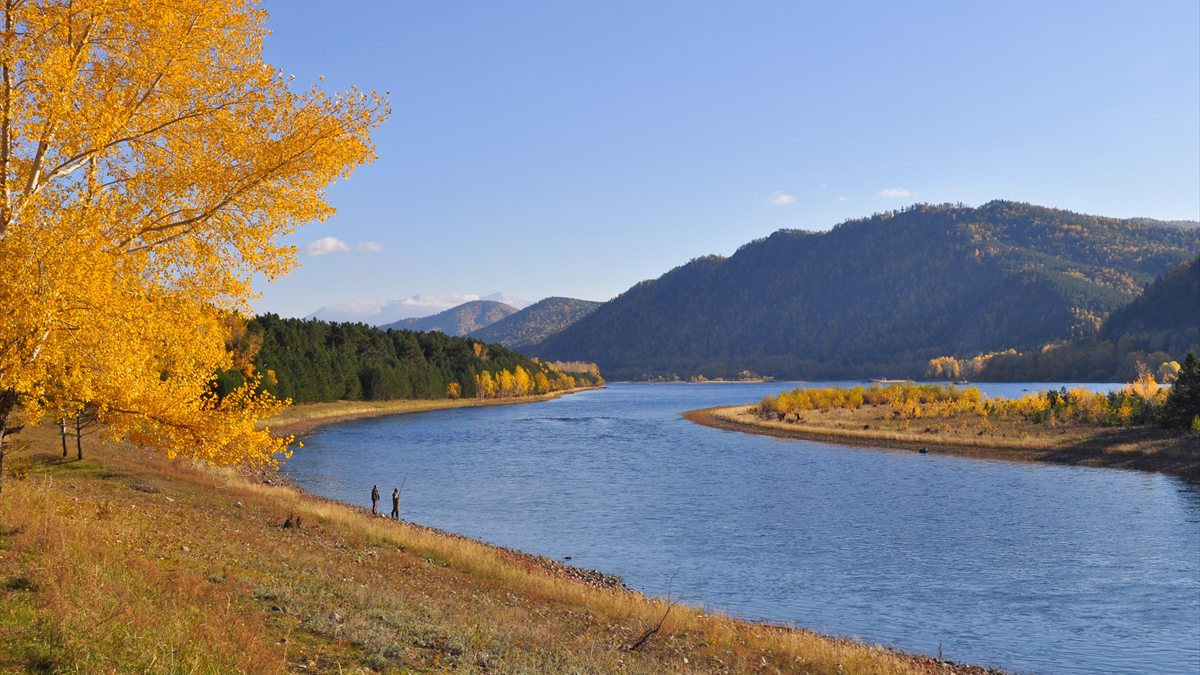}

            \tcblower

            \raggedright
            \small
            \textbf{Source:}
            He sat on the \amb{bank} of the river.

            \vspace{1mm}
            \textbf{Ambiguity:}
            \emph{bank} may denote a financial institution or a riverbank.

            \vspace{1mm}
            \textbf{Resolving evidence:}
            ``of the river'' selects the riverside sense.

            \vspace{1mm}
            \textbf{Visual role:}
            Not required.
        \end{examplepanel}
    \end{subfigure}
    \hfill
    \begin{subfigure}[t]{0.315\textwidth}
        \caption{\texttt{vision\_needed}}
        \label{fig:vision_needed_example}

        \begin{examplepanel}{visioncolor}
            \centering
            \includegraphics[
                width=\linewidth
            ]{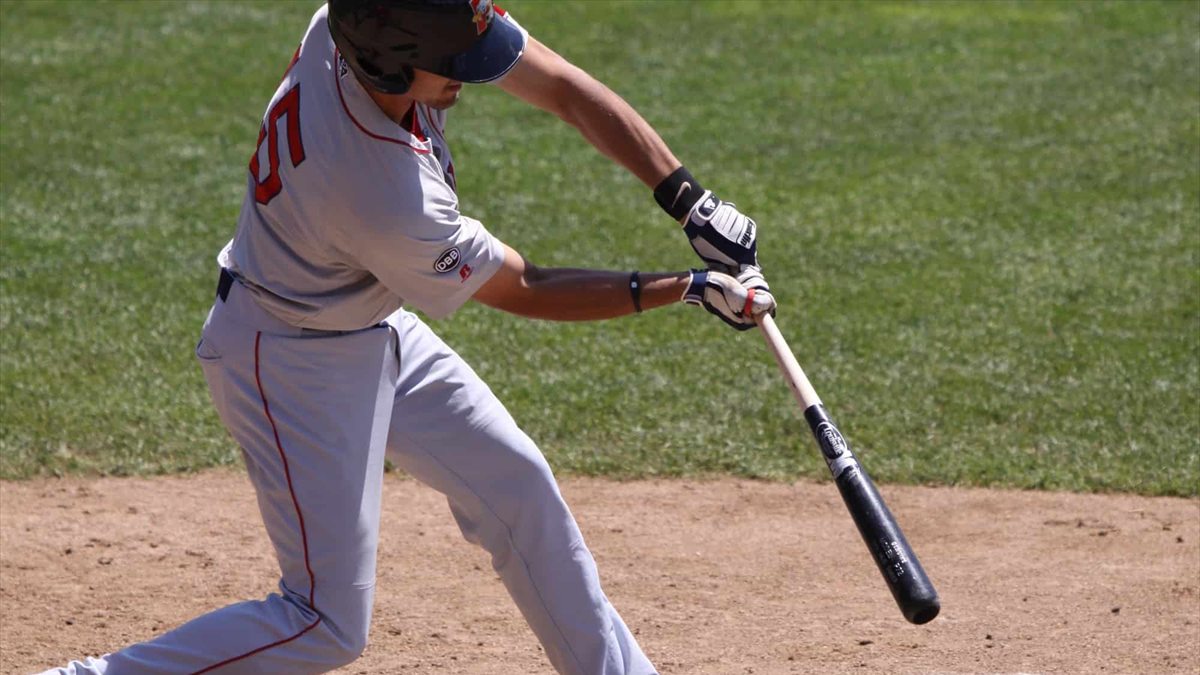}

            \tcblower

            \raggedright
            \small
            \textbf{Source:}
            He is holding a \amb{bat}.

            \vspace{1mm}
            \textbf{Ambiguity:}
            \emph{bat} may denote a baseball bat
            or the animal.

            \vspace{1mm}
            \textbf{Resolving evidence:}
            The image shows that the man is holding
            a baseball bat.

            \vspace{1mm}
            \textbf{Visual role:}
            Required.
        \end{examplepanel}
    \end{subfigure}

    \caption{
    Representative examples of the three visual-dependency labels.
    Highlighted spans indicate potentially ambiguous expressions.
    In \texttt{text\_resolved}, the textual context already determines
    the intended meaning, whereas \texttt{vision\_needed} requires
    discriminative evidence from the paired image.
    }
    \label{fig:label_examples}
\end{figure*}
Image-guided multimodal machine translation (MMT) incorporates aligned visual context to resolve translation-relevant uncertainty in the source sentence. Such uncertainty is especially common in short or underspecified descriptions, where multiple interpretations may lead to different target-language translations \citep{futeral_tackling_2023, ma_3am_2024}. Motivated by this, many prior studies have explored incorporating visual information into machine translation systems \citep{calixto_doubly-attentive_2017, libovicky_attention_2017}.

Despite evidence that visual contributions in MMT are highly sample-dependent—primarily beneficial when textual cues are insufficient \citep{elliott_adversarial_2018,caglayan_probing_2019}—prevailing systems still incorporate images unconditionally \citep{futeral_tackling_2023, futeral_towards_2025}. While recent adaptive approaches \citep{liu_detect_2025,guan_shift_2025,guan_dart_2026} introduce visual-input selection, their gating mechanisms are typically \textit{post-hoc} (i.e., triggered after visual encoding). Consequently, they fail to eliminate redundant computational overhead and potential visual interference. 

Motivated by these limitations, we propose an efficient, pre-visual routing mechanism.
This design is feasible because the usefulness of visual context is largely determined by whether translation-relevant uncertainty remains unresolved in the source sentence after accounting for its textual context \citep{caglayan_probing_2019}. The core challenge, then, lies in defining an accurate routing signal from the source. While a naive strategy might trigger visual invocation upon detecting any ambiguous word, lexical ambiguity does not equate to visual dependence. As shown in Figure~\ref{fig:label_examples}, the local ambiguity of "\emph{bank}" is fully resolved by the intra-sentence context ("of the river"), whereas "\emph{bat}" remains underdetermined without external visual evidence. This suggests that effective routing requires a sentence-level signal that distinguishes unresolved translation uncertainty from ambiguity already resolved by textual context.

In this work, we formulate visual invocation as a source-text sufficiency prediction problem: the source-only router determines whether the source sentence provides sufficient evidence for the relevant translation choice before any visual features are computed. If the text is insufficient, the paired image is invoked as external evidence; during data construction, we further verify that the image provides discriminative evidence for the unresolved interpretations.

Based on this formulation, we construct a visual-dependency dataset with three labels: \texttt{no\_ambiguity}, \texttt{text\_resolved}, and \texttt{vision\_needed}. We then train a lightweight XLM-RoBERTa-based \citep{conneau_unsupervised_2020} router that maps the first two categories to a text-only branch and \texttt{vision\_needed} to a multimodal branch. Since routing relies only on the source sentence and is independent of the translation backend, text-sufficient samples can bypass visual encoding and multimodal fusion entirely.

We evaluate the proposed router against rule-based, LLM-based, and learned baselines, and examine end-to-end translation under varying visual-call budgets. Our approach substantially reduces visual invocation while retaining most of the gains of always-on multimodal inference. Label-wise and case analyses further show that translation gains are concentrated in \texttt{vision\_needed} samples, highlighting the importance of distinguishing unresolved from text-resolved ambiguity.

Our main contributions are threefold:
\begin{itemize}
    \item We formulate selective visual use in MMT as a source-only, pre-visual routing problem based on source-text sufficiency, thereby distinguishing visual necessity from the mere presence of ambiguity.

    \item We construct a visual-dependency dataset with three fine-grained labels---\texttt{no\_ambiguity}, \texttt{text\_resolved}, and \texttt{vision\_needed}---and verify that the paired images of visually dependent samples contain discriminative evidence.

    \item We develop a lightweight, translation-backend-independent routing pipeline that substantially reduces visual processing while retaining most of the translation benefit provided by always-on multimodal inference.
\end{itemize}

%% file: pages/sections/2_relatedwork.tex
\section{Related Work}
\subsection{Multimodal Machine Translation and the Role of Vision}

Image-guided MMT uses aligned images to resolve linguistic uncertainty and enrich source representations. Multi30K established the standard image-caption translation setting \citep{elliott_multi30k_2016}. Early systems incorporated global or spatial features through visual and multi-source attention \citep{calixto_doubly-attentive_2017,libovicky_attention_2017}, while recent work adapts pretrained text-only translators or vision--language models \citep{futeral_tackling_2023,futeral_towards_2025}. Nevertheless, most systems still condition every sample on its image.

A parallel line of work has questioned whether MMT systems actually use this visual input. Adversarial image replacement or shuffling often has little effect on translation quality, suggesting weak visual grounding \citep{elliott_adversarial_2018}. Source-side masking experiments show that images become substantially more useful when textual evidence is removed, whereas intact captions are frequently sufficient by themselves \citep{caglayan_probing_2019}. Subsequent analyses attribute some apparent multimodal gains to regularisation rather than grounded visual use and show that conclusions about visual utility depend strongly on benchmark composition and image--text alignment \citep{wu_good_2021,li_vision_2021,long_exploring_2024}. \citet{li-etal-2022-vision} further show that stronger visual encoders can improve performance on text-masking probes, while gains on standard MMT benchmarks may not reliably reflect effective use of visual information. These studies establish that the contribution of vision is heterogeneous, but they primarily diagnose visual use at the model or dataset level. Our work instead turns this observation into a sample-level decision made before any visual features are computed.

\subsection{Ambiguity-Centric Evaluation and Visual-Dependency Data}

Several datasets target cases where images can affect translation. Multimodal Lexical Translation annotates ambiguous words and their target-language lexicalisations in visual and sentential context \citep{lala_multimodal_2018}. CoMMuTE pairs an ambiguous source sentence with contrastive images and translations corresponding to different senses, enabling targeted evaluation of visual disambiguation \citep{futeral_tackling_2023}. 3AM provides 26K English--Chinese image--sentence pairs selected using word-sense disambiguation signals \citep{ma_3am_2024}, while VIDA curates 2,500 visually dependent instances and introduces span-level, disambiguation-centric evaluation \citep{pan_vida_2026}.

These resources provide increasingly challenging tests of visual grounding, but they do not directly supply the mixed supervision required by a visual-invocation router. Multi30K mixes visually useful and text-sufficient samples without labeling them; 3AM identifies potential ambiguity but not whether sentential context resolves it; and CoMMuTE and VIDA emphasize visually dependent cases. Because ambiguity does not necessarily imply a need for vision, we distinguish \texttt{no\_ambiguity}, \texttt{text\_resolved}, and \texttt{vision\_needed}, verifying both textual insufficiency and discriminative evidence in the image. This taxonomy supplies positive and negative supervision for pre-visual routing.

\subsection{Selective Visual Use and Translation Routing}

DeDiT is the closest prior work: it detects contextually unresolved expressions and performs visual disambiguation only when needed \citep{liu_detect_2025}. However, because the image is already included in the vision--language model input, it does not avoid visual encoding. Similarly, SHIFT selects between text-only and key-frame inputs only after encoding candidate video frames \citep{guan_shift_2025}, while DART learns direct or cue-grounded translation within a unified video-conditioned model \citep{guan_dart_2026}. These methods adapt visual use after or within visual processing, whereas our text-only router makes the decision before the visual branch is invoked.

Related routing methods pursue different objectives. Using different model scales from the NiuTrans.LMT family as its small- and large-model translation backends \citep{luo-etal-2026-niutrans}, RouteLMT predicts the marginal gain of invoking the larger text-only translator and routes requests between model sizes under a fixed call budget \citep{luo_routelmt_2026}. VisNec, in contrast, estimates visual contribution through with- and without-image loss differences for multimodal data selection \citep{dong_visnec_2026}. Our work differs from both settings: rather than allocating requests between model capacities or selecting training data, we determine whether an additional visual modality should be invoked based on source-text sufficiency. This pre-visual, translation-backend-independent design allows text-sufficient inputs to bypass visual encoding and multimodal fusion entirely.

%% file: pages/sections/3_Methodology.tex
\section{Methodology}
\begin{figure*}[t]
    \centering
    \includegraphics[width=0.95\textwidth]{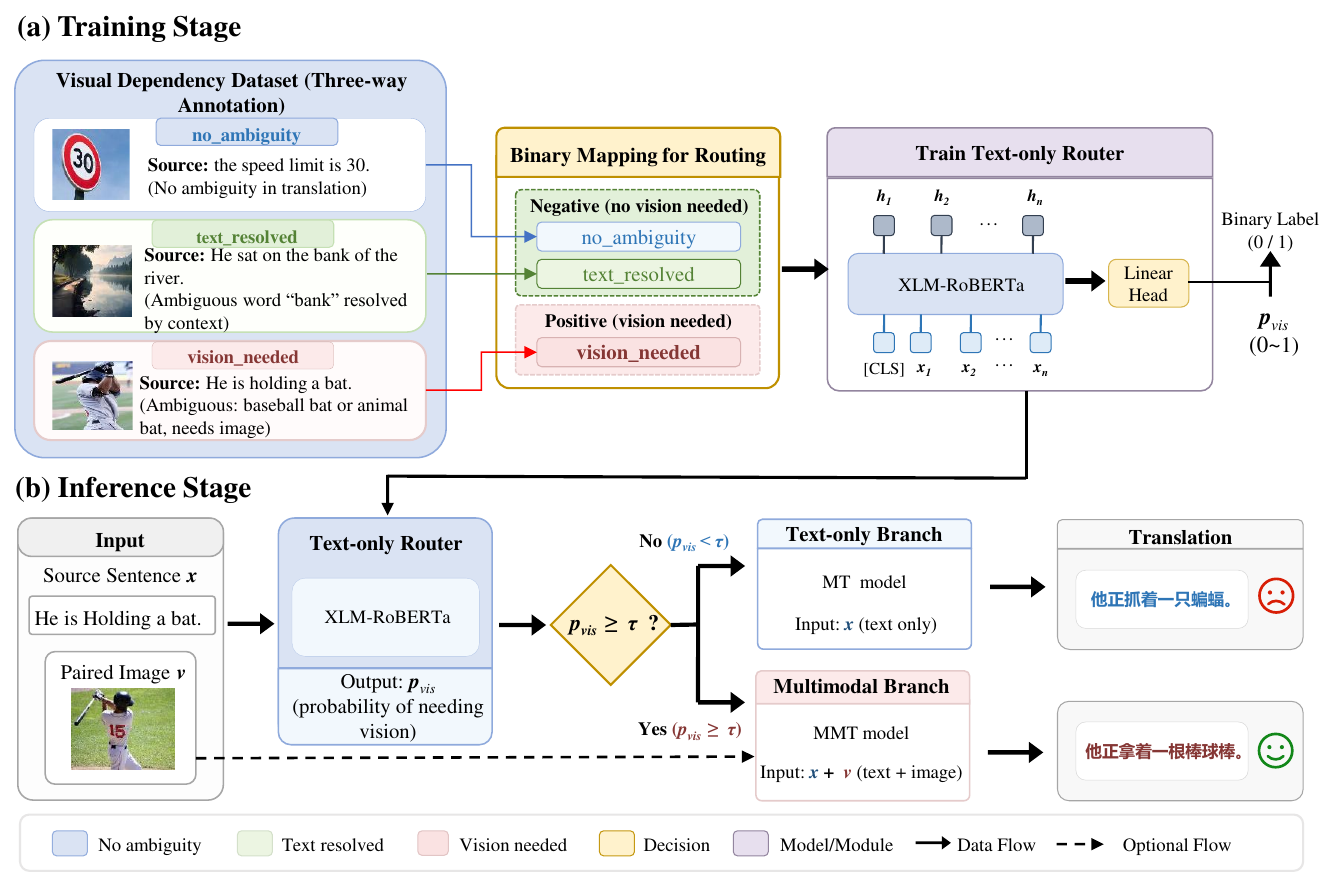}
   \caption{
Overview of the proposed pre-visual routing framework.
(a) A three-way visual dependency dataset is first constructed with the labels \texttt{no\_ambiguity}, \texttt{text\_resolved}, and \texttt{vision\_needed}. These annotations are then mapped to binary routing labels to train a text-only router based on XLM-RoBERTa. (b) At inference time, the router predicts $p_{\mathrm{vis}} = P(\mathrm{vision\_needed}\mid x)$ from the source sentence alone. The input is routed to the text-only branch when $p_{\mathrm{vis}} < \tau$, and to the multimodal branch otherwise. Both branches use the same translation backend and differ only in whether the paired image is provided.}
    \label{fig:method_overview}
\end{figure*}

To enable routing prior to visual encoding, we formally define the routing task and design a complete on-demand translation pipeline guided by a text-only router. As illustrated in Figure~\ref{fig:method_overview}, the three fine-grained visual-dependency labels are mapped to binary supervision for training the router. At inference time, the router predicts visual necessity from the source sentence alone, allowing text-sufficient samples to bypass visual encoding and multimodal fusion entirely.

\subsection{Problem Formulation}
Let $M$ be a translation backend supporting both text-only and image-text inputs. Each translation request consists of a source sentence $x$ and its paired image $v$. The system can utilize two branches of the backend: a computationally efficient text-only branch $M_t$ and a resource-intensive multimodal branch $M_m$. For a request $(x,v)$ these two branches generate translations as follows:
\begin{equation}
\hat{y}_{\mathrm{t}}
=
M_t(x),
\qquad
\hat{y}_{\mathrm{m}}
=
M_m(x,v)
\label{eq:two_translation_branches}
\end{equation}
The router selects which branch to use for serving the request, yielding the final output $\hat{y}$.

To determine whether the source sentence $x$ requires additional visual information before accessing the image, we design a pre-translation text-only router. Let $z(x)\in\{0,1\}$ denote the routing decision. Here $z=1$ indicates routing to $M_m$, meaning that relying solely on the source text is insufficient to support reliable translation, and the paired image may provide discriminative evidence for semantic choices. Conversely, $z=0$ indicates routing to $M_t$, meaning that the text information in the source sentence is already sufficient to support translation generation in the current direction.
 
We model the router's decision-making as a probabilistic prediction task. Given the input source sentence $x$, the text router predicts the probability of requiring visual context:
\begin{equation}
p_{\mathrm{vis}}
=
P_{\theta}(z=1\mid x)
\label{eq:visual_need_probability}
\end{equation}
where $\theta$ denotes the parameters of the router, and $p_{\mathrm{vis}}$ represents the probability that the source sentence has a visual demand.

Given a routing threshold $\tau$, the visual necessity probability is converted into a discrete routing decision:
\begin{equation}
\hat{z}=\mathbb{I}\left[p_{\mathrm{vis}}\geq\tau\right]
\label{eq:routing_decision}
\end{equation}
where $\mathbb{I}[\cdot]$ is the indicator function. When $p_{\mathrm{vis}}\geq\tau$, the system considers the current source sentence to have a high demand for vision and routes it to the multimodal translation branch; otherwise, the system deems the textual information sufficient and utilizes the text-only translation branch.

Accordingly, the entire on-demand multimodal translation process can be formalized as:
\begin{equation}
\hat{y}
=
\begin{cases}
M_t(x)
& \hat{z}=0,\\
M_m(x,v)
& \hat{z}=1,
\end{cases}
\label{eq:on_demand_translation}
\end{equation}

\subsection{Router}
\paragraph{Architecture.}
To predict the visual necessity $p_{\mathrm{vis}}$ defined in Eq.~\ref{eq:visual_need_probability}, we employ XLM-RoBERTa-base as a lightweight text-only router. Given a source sentence $x$, we extract the contextualized representation of its start token (e.g., \texttt{<s>}) to serve as the sentence-level embedding:
\begin{equation}
h_x
=
\operatorname{XLMR}_{\theta_e}(x)
\label{eq:router_representation}
\end{equation}
where $\theta_e$ denotes the parameters of the text encoder. Because $h_x$ is jointly encoded by the complete intra-sentence context, this representation can simultaneously capture potential ambiguous expressions and their contextual disambiguation cues. We then apply a linear classification head to map $h_x$ to the routing probability:
\begin{equation}
\boldsymbol{s}
=
Wh_x+b,
\qquad
p_{\mathrm{vis}}
=
\operatorname{softmax}(\boldsymbol{s})_1
\label{eq:router_prediction}
\end{equation}
where $W\in\mathbb{R}^{2\times d_h}$ and $b\in\mathbb{R}^{2}$ are learnable parameters, and the subscript $1$ indexes the \texttt{vision\_needed} class. This continuous probability is subsequently thresholded by $\tau$ to yield the discrete routing decision $\hat{z}$ (Eq.~\ref{eq:routing_decision}).
\paragraph{Training Objective.}
In practice, the distribution of visual dependency is inherently skewed, as genuine multimodal requirements are relatively sparse compared to text-sufficient samples. To prevent majority-class bias during training, we optimize the router using a class-weighted cross-entropy loss:
\begin{equation}
\mathcal{L} = -\frac{1}{N} \sum_{i=1}^{N} \sum_{c=0}^{1} w_c y_{i,c} \log P_{\theta}(z=c\mid x_i)
\label{eq:weighted_cross_entropy}
\end{equation}
where $y_{i,c}$ is the ground-truth routing label and $w_c = N / (2 N_c)$ is the inverse frequency weight for class $c$ ($N_c$ being the sample count of class $c$). 

We conduct full-parameter fine-tuning to jointly optimize the text encoder and the classification head. Guided by this balanced objective, the router learns to effectively weigh ambiguous expressions against their textual constraints to determine source-text sufficiency.

%% file: pages/sections/4_Dataset.tex
\section{Dataset}
To train and evaluate the proposed pre-visual router, we construct a visual-dependency dataset by re-annotating samples from 3AM, VIDA, and CoMMuTE \citep{ma_3am_2024,pan_vida_2026,futeral_tackling_2023}. The construction process combines translation-relevant ambiguity annotation, source-text sufficiency assessment, visual evidence verification, and human quality control. The following sections describe the label taxonomy, data construction procedure, and resulting dataset statistics.

\subsection{Label Taxonomy}
\label{subsec:label_taxonomy}
\begin{table}[t]
\centering
\small
\caption{Taxonomy of visual-dependency labels.}
\label{tab:label_taxonomy}

\renewcommand{\arraystretch}{1.05}

\begin{tabular*}{\columnwidth}{
    @{\extracolsep{\fill}}
    c c c c
    @{}
}
\toprule

\multirow{2}{*}{\textbf{Label}}
&
\multirow{2}{*}{\textbf{Ambiguity}}
&
\textbf{Text}
&
\textbf{Image}
\\[-1pt]

&
&
\textbf{sufficient}
&
\textbf{required}
\\

\midrule

\texttt{no\_ambiguity}
& No
& Yes
& No \\

\texttt{text\_resolved}
& Yes
& Yes
& No \\

\texttt{vision\_needed}
& Yes
& No
& Yes \\

\bottomrule
\end{tabular*}
\end{table}

As summarized in Table~\ref{tab:label_taxonomy}, we determine ground-truth visual dependency through three progressive criteria: potential translation ambiguity, textual sufficiency, and visual validity. This sequential verification yields three distinct categories. \textbf{\texttt{no\_ambiguity}} denotes sentences lacking any translation-relevant ambiguity. \textbf{\texttt{text\_resolved}} indicates that while local ambiguities exist, intra-sentence contextual cues fully determine the intended meaning. For both of these classes, image access is unnecessary. Conversely, \textbf{\texttt{vision\_needed}} signifies that the source text contains unresolved ambiguities, and the paired image explicitly provides the discriminative evidence required for accurate translation. Instances where both the text and image fail to resolve the ambiguity are discarded during data construction.

\subsection{Data Construction}
\label{subsec:data_construction}

\begin{figure}[t]
    \centering
    \includegraphics[
        width=\columnwidth,
        keepaspectratio
    ]{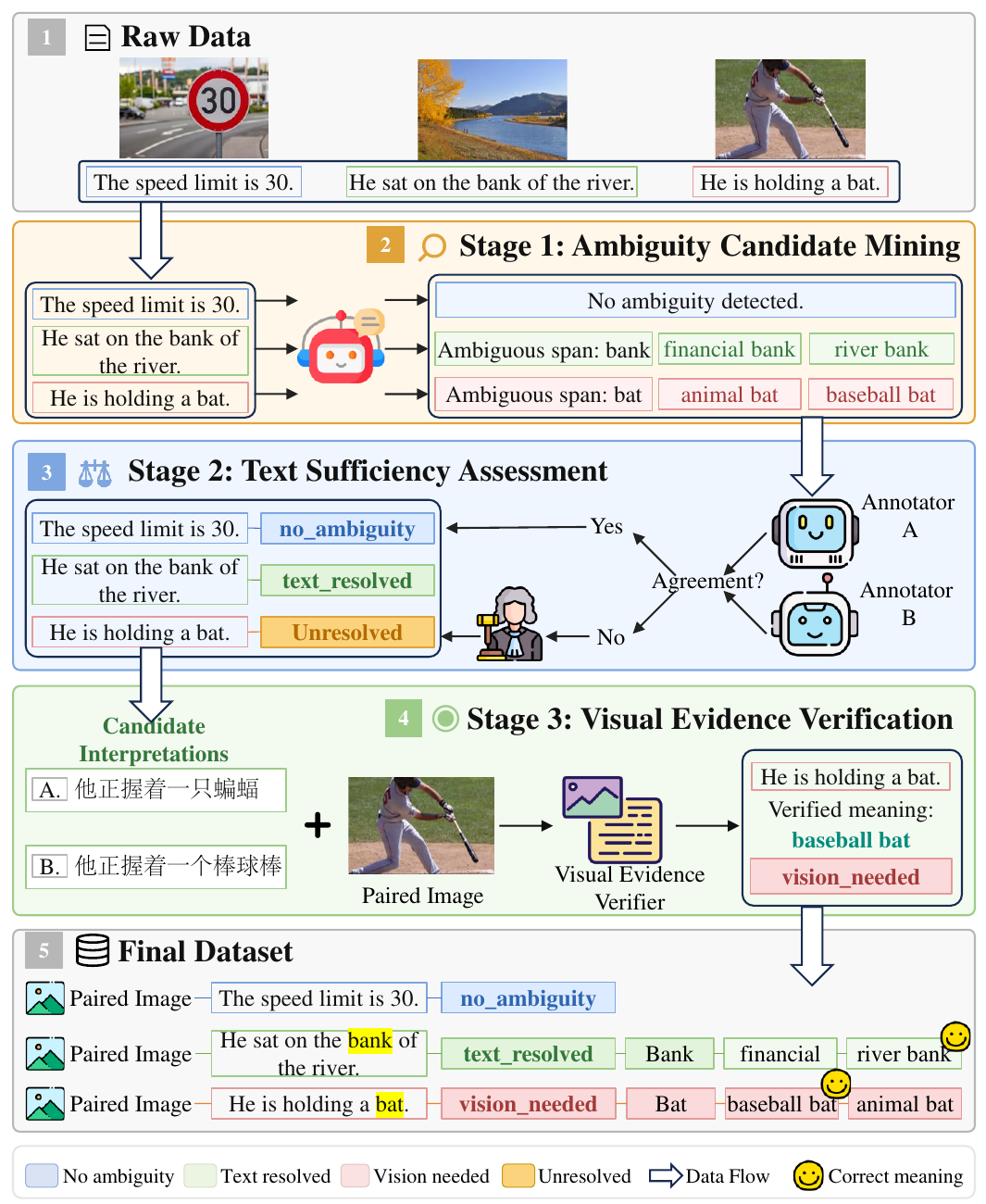}
    \caption{
    Overview of the visual-dependency data construction pipeline. Three examples illustrate how ambiguity candidate mining, text sufficiency assessment, and visual evidence verification produce the final labels: \texttt{no\_ambiguity}, \texttt{text\_resolved}, and \texttt{vision\_needed}. Only textually unresolved samples undergo visual verification.
    }
    \label{fig:data_construction}
\end{figure}

Figure~\ref{fig:data_construction} illustrates our
three-stage annotation pipeline, followed by human quality
control. Before annotation, we apply rule-based filtering
and source-sentence deduplication, then retain image--text
pairs only when the paired source sentence ranks within
the top 0.5\% of all texts for its image under SigLIP
similarity. Detailed filtering rules, adjudication criteria,
and quality-control statistics are provided in
Appendix~\ref{app:dataset_details}.

\paragraph{Ambiguity candidate mining.}
DeepSeek-V4-Flash extracts potentially ambiguous spans, ambiguity types, and candidate interpretations from each source sentence. We retain candidates whose plausible interpretations yield different target-language translations, and use spaCy part-of-speech analysis to filter out punctuation, function words, and unlikely ambiguity candidates. Figure~\ref{fig:data_construction} illustrates this step with \textit{bank} and \textit{bat}.

\paragraph{Text sufficiency assessment.}
Given the source sentence, candidate interpretations, and
target language, Qwen-Max and DeepSeek-V4-Pro independently
assign \texttt{no\_ambiguity}, \texttt{text\_resolved}, or
\texttt{unresolved}, without access to the paired image.
Their exact label agreement is 79.04\%. Agreements are
retained directly, while disagreements are adjudicated by
GPT-5.5 based on label validity, ambiguity interpretation,
textual evidence, and reasoning consistency. The top
annotation is accepted if its score is at least 11 out of
16 and leads by at least 2 points. Cases meeting the score
threshold but not the margin threshold undergo human
review; lower-scoring cases are discarded.

\paragraph{Visual evidence verification.}
For each \texttt{unresolved} sample, we convert the
candidate interpretations into complete descriptions
associated with distinct Chinese translations and use
SigLIP to score their compatibility with the paired image.
A sample is automatically labeled \texttt{vision\_needed}
when the top score is at least 0.60 on a $[0,1]$ scale and
its margin over the runner-up is at least 0.08. Cases
meeting the score threshold but not the margin threshold
undergo human review; those below the score threshold are
discarded. The two text-sufficient classes bypass this
stage.

\paragraph{Human quality control.}
Two computer science master's students, both native Chinese
speakers proficient in English, independently review
borderline cases and screen all 3,300 candidate instances
for label correctness, interpretation plausibility,
evidence sufficiency, and translation relevance. Their raw
pass/fail agreement before discussion is 77\%, with
disagreements resolved through discussion. Supported label
errors are corrected, while 247 invalid instances (7.48\%)
are removed, yielding the final 3,053-instance dataset.

\subsection{Dataset Statistics}
\label{subsec:dataset_statistics}

\begin{table}[t]
\centering
\small
\caption{Statistics of the visual-dependency dataset.
\texttt{no\_vision} combines \texttt{no\_ambiguity} and
\texttt{text\_resolved} as the negative class, while
\texttt{vision\_needed} is treated as the positive class.}
\label{tab:dataset_statistics}

\setlength{\tabcolsep}{3pt}
\renewcommand{\arraystretch}{1.05}

\resizebox{\columnwidth}{!}{
\begin{tabular}{lrrrrr}
\toprule
Split
& No amb.
& Text res.
& No vis.
& Vis. req.
& Total \\
\midrule
Train & 832   & 616 & 1,448 & 973   & 2,421 \\
Dev   & 104   & 77  & 181   & 107   & 288   \\
Test  & 104   & 78  & 182   & 162   & 344   \\
\midrule
Total & 1,040 & 771 & 1,811 & 1,242 & 3,053 \\
\bottomrule
\end{tabular}
}

\end{table}

The dataset is constructed primarily from 3AM and VIDA, with deduplicated and re-annotated CoMMuTE instances included to evaluate generalisation to an independent data source. Exact and near-duplicate removal is performed before data splitting to prevent overlap across partitions. After annotation and quality control, the final dataset contains 3,053 instances. The label distribution across the training, development, and test sets is reported in Table~\ref{tab:dataset_statistics}.

%% file: pages/sections/5_Experiment.tex
\section{Experiments}
\label{sec:experiments}

\subsection{Data and Model}
\label{subsec:setup}

\paragraph{Data}
To adapt the dataset for the binary routing task, we group \texttt{no\_ambiguity} and \texttt{text\_resolved} into a single negative class (\textit{vision-unnecessary}), retaining \texttt{vision\_needed} as the positive class (\textit{vision-required}). As detailed in Table~\ref{tab:dataset_statistics}, the dataset exhibits a mild class imbalance (roughly a 2:3 positive-to-negative ratio), which we address by applying inverse-frequency class weights derived from the training set. The test split is then used for a dual evaluation: assessing the standalone classification accuracy of the router and measuring the end-to-end performance of the on-demand translation pipeline.

\paragraph{Model}

To prevent different model architectures from introducing confounding variables into the analysis of visual contributions, our adaptive framework employs a unified Vision-Language Model (VLM) backend. The text-only and multimodal branches strictly share the same parameters, instructions, and decoding configurations; the only difference is the presence or absence of visual input.

\begin{figure*}[t]
    \centering
    \includegraphics[width=1.0\textwidth]
    {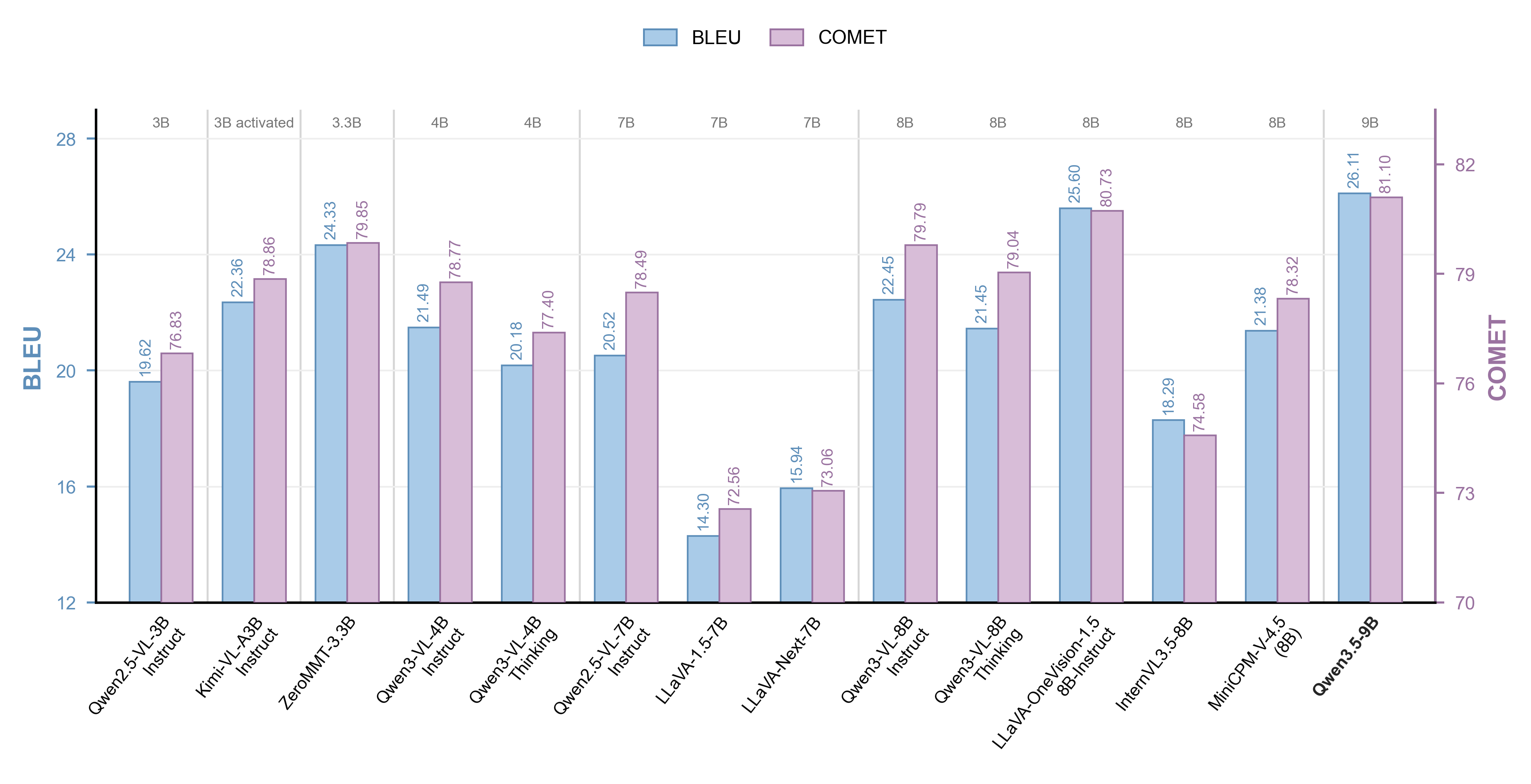}
    \caption{
    Translation performance of candidate vision-language models
    on Multi30K, macro-averaged over the
    En$\rightarrow$De, En$\rightarrow$Fr, and
    En$\rightarrow$Cs directions. Models are ordered by their advertised parameter
    scales.
    }
    \label{fig:model_comparison}
\end{figure*}

Under a unified zero-shot evaluation on Multi30K,
Qwen3.5-9B achieves the best macro-average performance
among the candidate VLMs (26.11 BLEU and 81.10 COMET;
Figure~\ref{fig:model_comparison}). We therefore use it
as the shared backend for both translation branches.

\subsection{Baselines}
\label{subsec:translation_strategies}
To comprehensively evaluate the effectiveness of our method, we establish two sets of evaluation settings: routing baselines to assess the standalone classification accuracy of our text-only router, and end-to-end translation strategies to demonstrate the efficacy of the entire on-demand pipeline.

\paragraph{Routing Baselines}
To evaluate the classification performance of our fully fine-tuned text-sufficiency router (based on XLM-RoBERTa), we compare it against the following baselines:
\begin{itemize}
\item \textbf{Lexicon-based Heuristic}: A rule-based baseline that triggers the multimodal branch if the source sentence contains words from a predefined ambiguous lexicon; otherwise, it relies solely on text.
\item \textbf{LLM Zero-Shot}: A prompting-based baseline where a state-of-the-art LLM (DeepSeek-V4-Pro) is instructed to determine zero-shot whether the current sentence requires visual supplementation based on textual semantics.
\item \textbf{Frozen Encoder + Linear}: An ablation baseline that freezes the XLM-RoBERTa backbone and only trains the linear classification head, evaluating the necessity of full-parameter fine-tuning.
\end{itemize}

\paragraph{End-to-End Translation Baseline}
We compare PreVIR with endpoint policies, budget-matched routing baselines, and oracle references. \textbf{Text-only} and \textbf{Always-Multimodal} route all samples to the text-only and multimodal branches, respectively. For a fair comparison, \textbf{Random}, \textbf{Ambiguity-based}, \textbf{Length-based}, \textbf{Sentinel-src-25}, and \textbf{PreVIR} operate under the same 46.22\% visual budget. Random routing selects samples uniformly; Ambiguity-based routing prioritizes sentences containing more lexical ambiguity cues; Length-based routing prioritizes shorter sentences with less textual context; and Sentinel-src-25 \citep{proietti-etal-2025-estimating} prioritizes sentences with lower predicted translation quality. PreVIR ranks samples by their predicted visual-necessity probabilities. Finally, the \textbf{Gold-label Router} uses the annotated visual-dependency labels, while the \textbf{Gain-Oracle} selects the multimodal output only when it yields a positive reference-aware COMET gain.

\subsection{Implementation Details}
\label{subsec:implementation}

\paragraph{Router Training.}
We fully fine-tune XLM-RoBERTa-base for 20 epochs on a single NVIDIA A800 80GB GPU. To strictly prevent information leakage, the router accesses only the source text during both training and inference. We optimize the model using AdamW with a learning rate of $2\times10^{-5}$ and a batch size of 8. To mitigate class imbalance in our training data, we employ a weighted cross-entropy loss with inverse-frequency weights (0.836 for the negative class and 1.244 for the positive \texttt{vision\_needed} class). At inference, the threshold $\tau$ is set to 0.5.

\paragraph{Translation Backend.}%
We deploy Qwen3.5-9B using vLLM \citep{kwon_efficient_2023} in bfloat16 precision. For both branches, we apply deterministic greedy decoding (temperature = 0, top-p=1.0). All generated outputs undergo a post-processing step to strip non-translation artifacts such as code blocks, thinking traces, and prefixes.

%% file: pages/sections/6_Result_and_Analysis.tex
\section{Results and Analysis}

\subsection{Routing Results}
\begin{table}[t]
\centering
\small
\renewcommand{\arraystretch}{1.06}
\caption{
Visual-dependency routing results on the test set. ZS denotes zero-shot inference. P, R, and F1 denote precision, recall, and F1-score, respectively, for the positive \texttt{vision\_needed} class. Call denotes the percentage of samples routed to the multimodal translation path.}
\label{tab:router_results}

\begin{tabular*}{\columnwidth}{
@{\extracolsep{\fill}}
l c c c c c
@{}
}
\toprule
Method
& Acc.$\uparrow$
& P$\uparrow$
& R$\uparrow$
& F1$\uparrow$
& Call (\%) \\
\midrule

Always Neg.
& 0.53
& 0.00
& 0.00
& 0.00
& 0.00 \\

Lexicon Rule
& 0.44
& 0.42
& 0.48
& 0.45
& 53.78 \\

DeepSeek-ZS
& \underline{0.84}
& \textbf{0.91}
& 0.73
& \underline{0.81}
& 38.08 \\

Frozen XLM-R
& 0.47
& 0.47
& \textbf{1.00}
& 0.64
& 100.00 \\

\midrule

\textbf{PreVIR (Ours)}
& \textbf{0.86}
& \underline{0.86}
& \underline{0.85}
& \textbf{0.85}
& 46.22 \\

\bottomrule
\end{tabular*}

\end{table}

Table~\ref{tab:router_results} reports routing performance at the default threshold $\tau=0.5$. A useful pre-visual router must balance two competing objectives: covering \texttt{vision\_needed} instances so that potentially useful visual evidence is not missed, while avoiding unnecessary visual calls for text-sufficient inputs. The two extreme baselines illustrate why no single metric is sufficient for evaluating this trade-off. Always Neg.\ benefits from the negative-class majority but fails to identify any visually dependent sample, whereas Frozen XLM-R attains perfect recall only by routing every input to the multimodal branch. Among the non-degenerate strategies, PreVIR achieves the best overall accuracy and F1 score. DeepSeek-ZS maintains the highest precision but follows a more conservative routing policy, missing more than one quarter of the \texttt{vision\_needed} instances. PreVIR improves their coverage while retaining selective visual invocation, yielding a stronger overall balance between missed visual evidence and unnecessary calls.

\begin{figure}[t]
    \centering
    \includegraphics[width=\columnwidth]{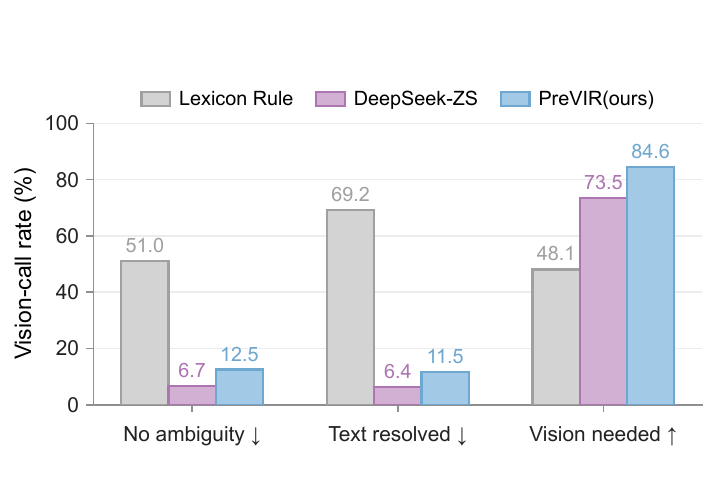}
    \caption{
    Vision-call rates across the three fine-grained visual-dependency
    categories. Lower rates are preferred for
    \texttt{no\_ambiguity} and \texttt{text\_resolved}, while a higher
    rate is preferred for \texttt{vision\_needed}.
    }
    \label{fig:labelwise_call_rate}
\end{figure}

The category-wise results in Figure~\ref{fig:labelwise_call_rate} explain where this advantage comes from. The Lexicon Rule invokes vision more frequently for \texttt{text\_resolved} samples than for \texttt{vision\_needed} samples (69.2\% versus 48.1\%), showing that the presence of a potentially ambiguous expression is poorly aligned with visual dependency when the surrounding text has already resolved its meaning. In contrast, PreVIR invokes vision for only 11.5\% of the \texttt{text\_resolved} samples while covering 84.6\% of the \texttt{vision\_needed} samples. This separation is consistent with the router using sentence-level textual context rather than treating ambiguity itself as a visual trigger. 

Overall, the results support source-text sufficiency as an effective signal for pre-visual routing: task-specific supervision does not simply increase the number of visual calls, but directs them toward samples whose translation-relevant uncertainty remains unresolved by text. The remaining routing errors also indicate room for improving the identification of difficult visual-dependency cases, whose downstream translation effects are examined in the following experiments.

\subsection{End-to-End Translation Results}

\begin{table*}[t]
\centering
\small
\renewcommand{\arraystretch}{1.08}

\caption{
Overall and label-wise translation results. No amb., Text res., and Vision req. correspond to the \texttt{no\_ambiguity}, \texttt{text\_resolved}, and \texttt{vision\_needed} gold labels, respectively. Random, Ambiguity-based, Length-based, Sentinel-src-25, and PreVIR are evaluated under the same 46.22\% vision budget. Text-only and Always-Multimodal are endpoint policies, while the oracle methods are diagnostic references evaluated at their native call rates. Bold and underlined values denote the best and second-best results, respectively, among the budget-matched routing methods.
}
\label{tab:translation_results}

\resizebox{\textwidth}{!}{%
\begin{tabular}{@{}lrrrrrrrrr@{}}

\toprule
& \multicolumn{4}{c}{BLEU $\uparrow$}
& \multicolumn{4}{c}{COMET $\uparrow$}
& \multicolumn{1}{c}{} \\

\cmidrule(lr){2-5}
\cmidrule(lr){6-9}

Strategy
& Overall
& No amb.
& Text res.
& Vision req.
& Overall
& No amb.
& Text res.
& Vision req.
& Vision Call\\

\midrule
Gain-Oracle
& 44.83
& 49.66
& 45.57
& 38.05
& 88.93
& 90.70
& 88.56
& 87.96
& 41.57\% \\

Gold-label Router
& 41.96
& --
& --
& --
& 87.66
& --
& --
& --
& 47.09\% \\

\midrule
Text-only
& 40.56
& 46.24
& 42.88
& 31.19
& 85.75
& 89.19
& 87.31
& 82.79
& 0.00\% \\

Always-Multimodal
& 41.93
& 45.65
& 43.27
& 35.73
& 87.79
& 89.51
& 87.45
& 86.85
& 100.00\% \\

\midrule
Random
& 41.19
& 45.98
& \underline{43.03}
& 33.29
& 86.69
& 89.35
& 87.37
& 84.66
& 46.22\% \\

Ambiguity-based
& \textbf{42.03}
& \textbf{47.51}
& \textbf{43.08}
& \underline{34.12}
& \underline{87.04}
& \textbf{89.56}
& 87.45
& \underline{85.23}
& 46.22\% \\

Length-based
& 41.41
& 47.11
& 42.98
& 32.77
& 86.11
& 89.44
& 87.33
& 83.39
& 46.22\% \\

Sentinel-src-25
& \underline{41.63}
& \underline{47.27}
& 42.80
& 33.16
& 86.83
& \underline{89.49}
& \textbf{87.58}
& 84.76
& 46.22\% \\

\textbf{PreVIR (Ours)}
& 41.53
& 46.36
& 41.36
& \textbf{35.82}
& \textbf{87.47}
& 89.31
& \underline{87.55}
& \textbf{86.26}
& 46.22\% \\

\bottomrule

\end{tabular}
}

\end{table*}

Table~\ref{tab:translation_results} compares endpoint policies, budget-matched routing strategies, and oracle diagnostics in terms of translation quality and visual invocation rate using Qwen3.5-9B as the translation backend. Results for additional backends are reported in Appendix~\ref{app:cross_backend}. 

\paragraph{Budget-Matched Comparison.} Under the same 46.22\% visual budget, PreVIR achieves the highest overall COMET score. Although the Ambiguity-based strategy obtains the best overall BLEU, PreVIR performs best on the \texttt{vision\_needed} subset in both metrics, outperforming it by 1.70 BLEU and 1.03 COMET. Meanwhile, COMET differences on the two text-sufficient subsets remain within 0.25 points. These results indicate that PreVIR allocates more of the fixed visual budget to genuinely vision-dependent inputs.

\paragraph{Quality--Efficiency Trade-off.}
Compared with Text-only translation, PreVIR improves BLEU by 0.97 and COMET by 1.72. It remains within 0.40 BLEU and 0.32 COMET of Always-Multimodal while reducing visual invocation by 53.78\%. Paired significance tests show that the COMET improvement over Text-only is statistically significant, whereas the BLEU improvement is not. Neither metric differs significantly from Always-Multimodal. Detailed confidence intervals and significance tests are reported in Appendix~\ref{app:reliability}. Therefore, selective routing recovers most of the benefit of visual context with fewer than half of the visual calls. At matched request batch sizes, PreVIR also improves throughput by 18.7--46.1\% over Always-Multimodal, including router overhead. Detailed settings and timing results are provided in Appendix~\ref{app:efficiency}.

\paragraph{Oracle Diagnostics.} The Gold-label Router exceeds PreVIR by only 0.43 BLEU and 0.19 COMET, showing that relatively little performance is lost due to routing prediction errors. The Gain-Oracle achieves substantially higher performance by selecting outputs according to their realized reference-aware gains. This gap suggests that annotated visual necessity and actual backend gain are related but not identical, as the latter also depends on image informativeness and the backend's ability to exploit visual evidence.

\subsection{Quality--Budget Trade-off}
\label{subsec:threshold_analysis}

\begin{figure}[t]
    \centering
    \includegraphics[width=\columnwidth]{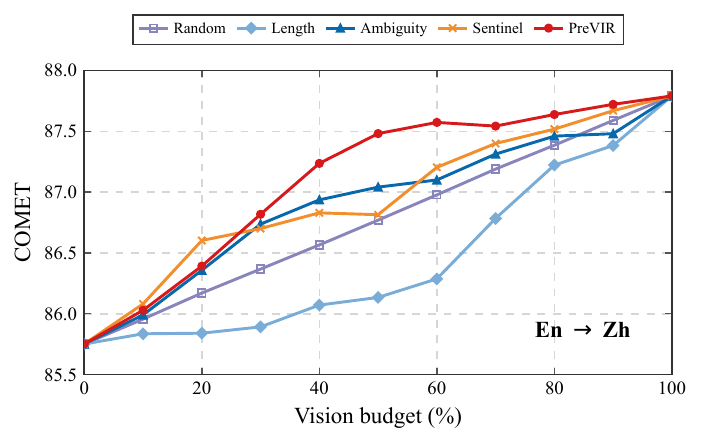}
    \caption{
    Quality–-budget trade-offs of hybrid translation routing.
    }
    \label{fig:comet_buget_trade-off}
\end{figure}

We further explored the performance of various methods at different visual invocation rates. As shown in Figure~\ref{fig:comet_buget_trade-off}, Sentinel-src-25 performs better under extremely restricted budgets of 10\% and 20\%. PreVIR becomes the strongest method at 30\% and achieves the highest COMET at every evaluated budget from 30\% to 90\%, with its clearest advantage appearing in the practical 40--60\% range. This trend shows that the improvement observed at the default 46.22\% operating point extends across a broad range of vision budgets. 

The relative performance of the heuristic methods further illustrates the importance of the routing signal. Length-based routing remains below random routing at every intermediate budget, indicating that sentence length provides an ineffective prioritization signal in this setting. Ambiguity-based routing performs better than random at most budgets, suggesting that lexical ambiguity contains useful coarse information. Nevertheless, it remains below PreVIR throughout the 10--90\% range, particularly at intermediate budgets. This consistent gap supports the distinction between detecting potentially ambiguous expressions and determining whether the corresponding uncertainty remains unresolved by the textual context.

The curve also reveals a limitation of necessity-based routing. Although PreVIR is optimized to identify \texttt{vision\_needed} samples, Sentinel-src-25 performs better when only the top 10--20\% of inputs can invoke vision. This suggests that identifying whether vision is needed is not identical to ranking the magnitude of the translation gain realized by a particular multimodal backend. By the 50\% budget point, PreVIR has already selected 143 of the 162 gold \texttt{vision\_needed} samples and retains approximately 84.8\% of the full COMET improvement delivered by Always-Multimodal inference. Consequently, the curve largely plateaus beyond this point, with minor non-monotonic fluctuations. These results characterize PreVIR as an effective visual-necessity router while leaving backend-specific gain prediction as a complementary direction for future work.

\begin{CJK*}{UTF8}{gbsn}
\subsection{Case Studies}
\label{subsec:case_study}

\begin{table*}[t]
\centering
\scriptsize
\setlength{\tabcolsep}{3.2pt}
\renewcommand{\arraystretch}{1.08}

\caption{
Case studies of pre-visual routing. Highlighted spans indicate translation-relevant ambiguity or its corresponding realization in the translation. COMET scores are reported for the two translation branches.}
\label{tab:case_study}

\begin{tabularx}{\textwidth}{
@{}
p{0.075\textwidth}
p{0.13\textwidth}
p{0.072\textwidth}
X
p{0.055\textwidth}
@{}
}
\toprule
Case
& Image
& Item
& Text
& COMET \\
\midrule
\multirow{4}{0.075\textwidth}{
\centering
Multimodal Branch\\
Vision req.\\
Δ=+40.308
}
&
\multirow{4}{0.13\textwidth}{
\centering
\includegraphics[
    height=1.2cm, 
    keepaspectratio
]{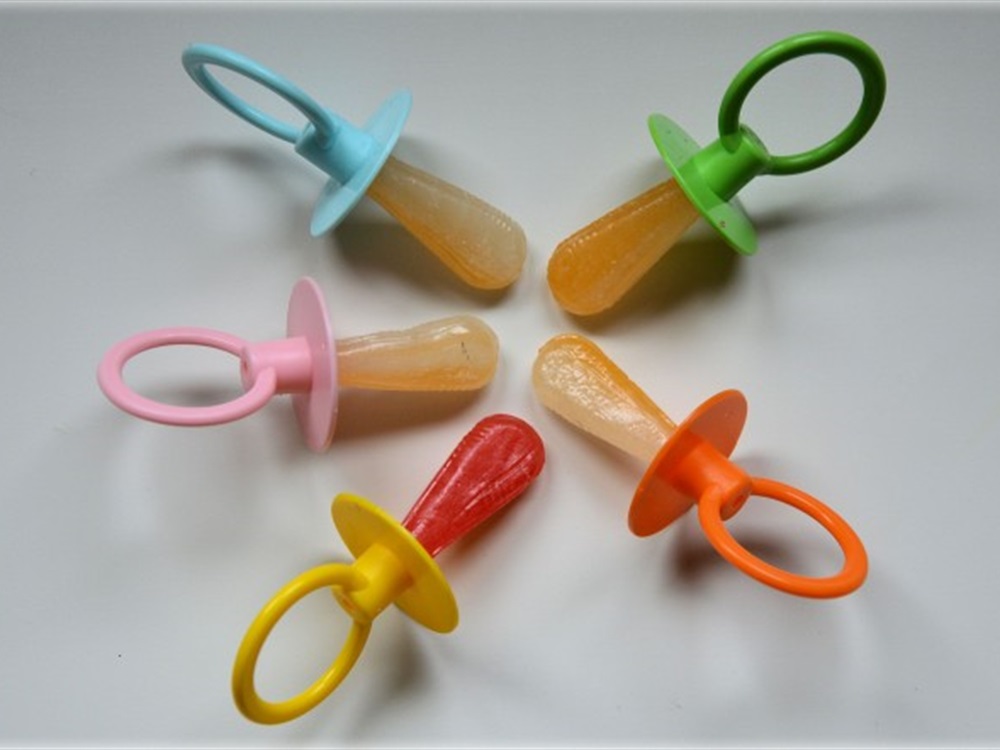}
}
&
Source
&
The \colorbox{gray!15}{dummies} need to be cleaned.
&
-- \\

&&
Reference
&
\colorbox{gray!15}{奶嘴}需要清洗。
&
-- \\

&&
Text-only
&
\colorbox{gray!15}{假人}需要被清理。
&
57.642 \\

&&
Multimodal
&
\colorbox{gray!15}{奶嘴}需要清洗。
&
97.950 \\

\midrule
\multirow{4}{0.075\textwidth}{
\centering
Text-only Branch\\
Text res.\\
Δ=0.000
}
&
\multirow{4}{0.13\textwidth}{
\centering
\includegraphics[
    height=1.2cm,
    keepaspectratio
]{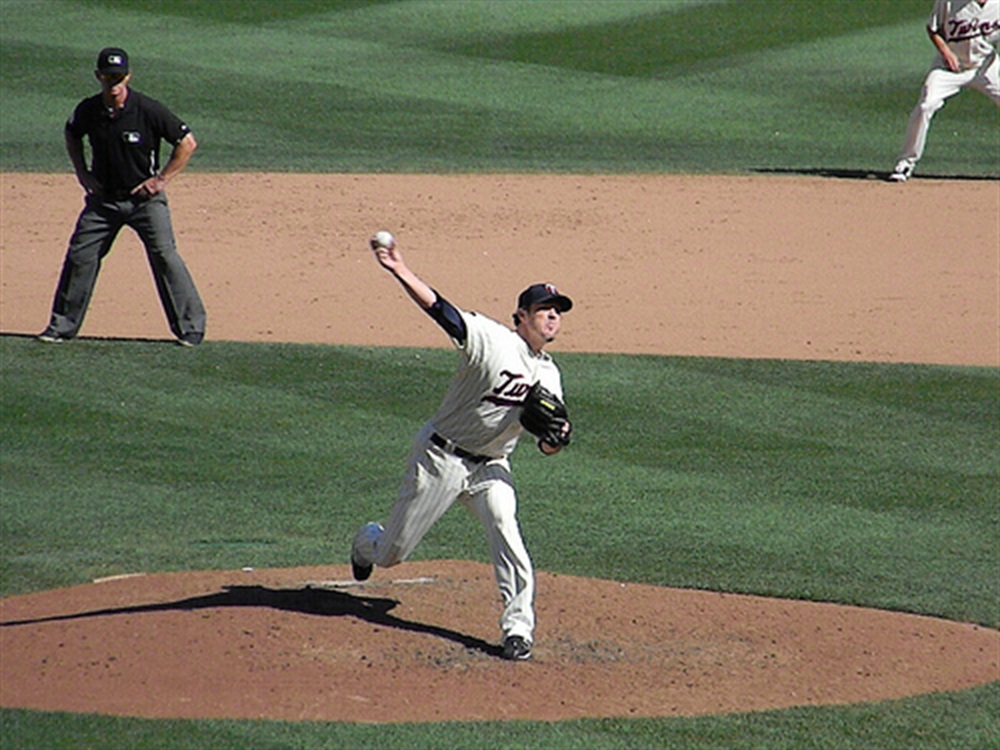}
}
&
Source
&
A \textbf{baseball} \colorbox{gray!15}{pitcher}
is throwing a \colorbox{gray!15}{pitch}.
&
-- \\

&&
Reference
&
一个棒球\colorbox{gray!15}{投手}正在投出一个\colorbox{gray!15}{球}。
&
-- \\

&&
Text-only
&
一名棒球\colorbox{gray!15}{投手}正在投\colorbox{gray!15}{球}。
&
92.267 \\

&&
Multimodal
&
一名棒球\colorbox{gray!15}{投手}正在投\colorbox{gray!15}{球}。
&
92.267 \\

\midrule
\multirow{4}{0.075\textwidth}{
\centering
Text-only Branch\\
Vision req.\\
Δ=+30.650
}
&
\multirow{4}{0.13\textwidth}{
\centering
\includegraphics[
    height=1.2cm,
    keepaspectratio
]{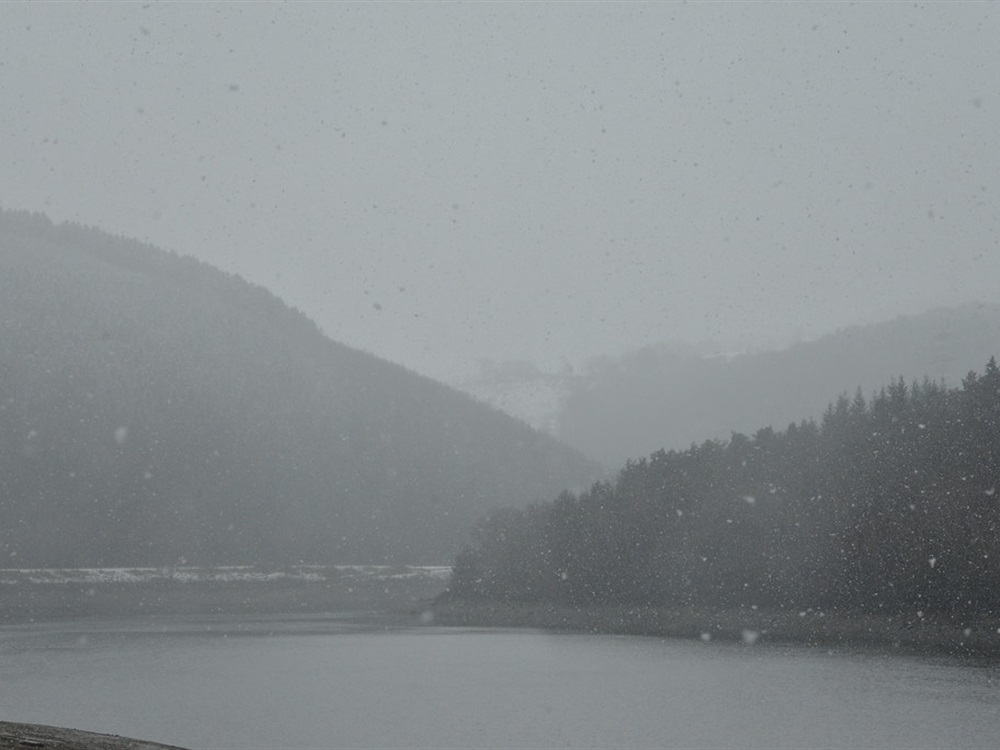}
}
&
Source
&
I know -- good job there are
\colorbox{gray!15}{showers}.
&
-- \\

&&
Reference
&
我知道，幸运的是下了\colorbox{gray!15}{阵雨}。
&
-- \\

&&
Text-only
&
我知道——太好了，这里有\colorbox{gray!15}{淋浴间}。
&
55.420 \\

&&
Multimodal
&
我知道——幸好有\colorbox{gray!15}{阵雨}。
&
86.070 \\

\midrule
\multirow{4}{0.075\textwidth}{
\centering
Multimodal Branch\\
Text res.\\
Δ=-3.289
}
&
\multirow{4}{0.13\textwidth}{
\centering
\includegraphics[
    height=1.2cm,
    keepaspectratio
]{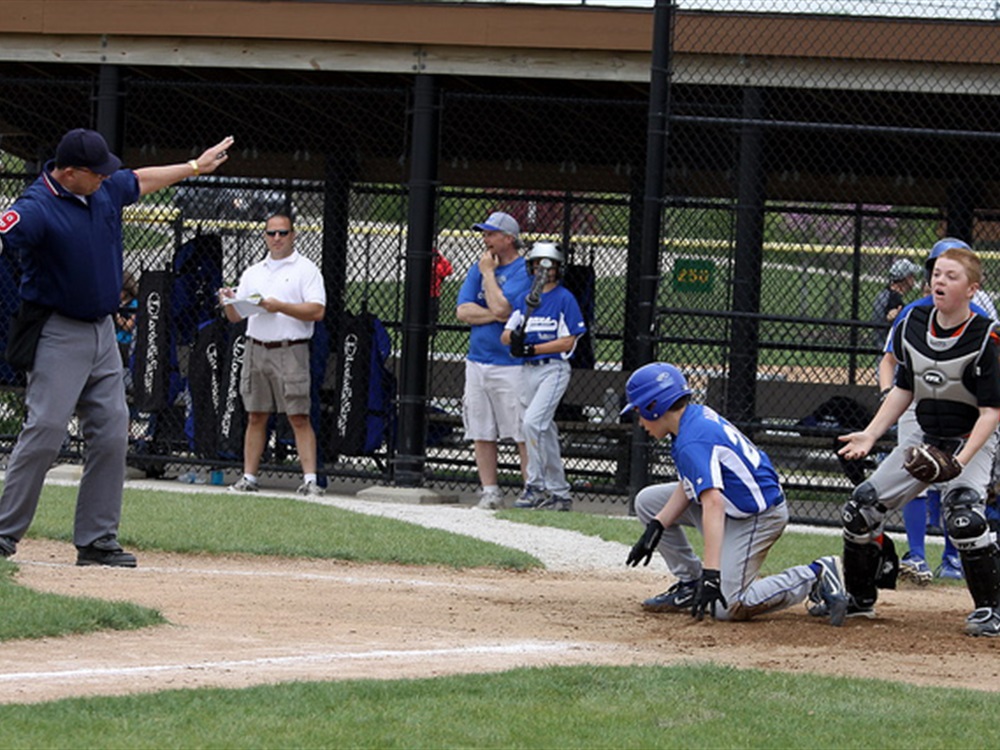}
}
&
Source
&
A baseball player slides into \textbf{home plate}
and is \colorbox{gray!15}{called safe} by the umpire.
&
-- \\

&&
Reference
&
一名棒球运动员滑进本垒，被裁判\colorbox{gray!15}{判为安全}。
&
-- \\

&&
Text-only
&
一名棒球运动员滑垒至本垒，裁判\colorbox{gray!15}{判其安全}。
&
87.958 \\

&&
Multimodal
&
一名棒球选手滑垒至本垒，裁判\colorbox{gray!15}{判定其安全上垒}。
&
84.669 \\

\bottomrule
\end{tabularx}
\end{table*}

As shown in Table~\ref{tab:case_study}, the successful routing cases (TP and TN) demonstrate PreVIR's ability to distinguish unresolved visual dependency from surface-level lexical ambiguity. Our router correctly invokes the multimodal branch when the source text lacks cues to disambiguate terms like \emph{dummies}. Conversely, it accurately bypasses visual processing when intra-sentence context (e.g., \emph{baseball}) already resolves local ambiguities like \emph{pitcher} and \emph{pitch}, thereby eliminating redundant computation.

Furthermore, the error cases (FN and FP) highlight a critical asymmetric risk in multimodal routing. Under-invocation (FN) on genuinely vision-dependent instances (e.g., \emph{showers}) deprives the model of necessary evidence, leading to severe adequacy drops. In contrast, over-invocation (FP) on text-sufficient instances primarily wastes the computational budget and introduces minor quality fluctuations without reliable gains. This asymmetry underscores that maintaining high recall on truly vision-required samples is paramount for robust on-demand MMT.

\end{CJK*}

%% file: pages/sections/7_Conclusion.tex
\section{Conclusion}
In this paper, we formulate visual invocation in MMT as source-text sufficiency prediction and propose PreVIR, a lightweight, translation-backend-independent router. By distinguishing genuine visual necessity from contextually text-resolved ambiguity, we construct a fine-grained dataset to supervise this pre-visual routing. Experiments demonstrate that PreVIR establishes a highly favorable performance-efficiency trade-off, preserving the vast majority of multimodal translation gains while substantially reducing visual calls. Label-wise, budgeted, and case analyses confirm that PreVIR succeeds by strictly allocating visual computation to genuinely vision-dependent samples. Ultimately, our findings establish source-only pre-visual routing as a highly efficient and principled alternative to indiscriminate multimodal processing.

%% file: pages/sections/8_Limitations.tex
\section*{Limitations}
Our study presents three main limitations. First, our 3,053-instance dataset focuses on English-to-Chinese translation. Because visual necessity inherently depends on target-language lexicalization, this scope may not fully capture diverse cross-lingual ambiguities. Second, strictly pre-visual routing prevents the model from identifying uninformative or misaligned images at test time, which poses risks when deploying on uncurated, noisy corpora. Third, the router optimizes for binary visual necessity rather than realized translation gain. While effective at general classification, it struggles to optimally rank high-gain instances under extremely tight visual budgets. Future work will explore multilingual expansion and gain-aware ranking objectives.

\section*{Acknowledgments}
This work was supported in part by the National Science Foundation of China (Nos. 62276056 and U24A20334), the Yunnan Fundamental Research Projects (No.202401BC070021), the Yunnan Science and Technology Major Project (No. 202502AD080014), the Liaoning Provincial Science and Technology Plan Project (No. 2026JH40/10100033), the Fundamental Research Funds for the Central Universities (Nos. N26XQD028, N25BSS054 and N25BSS094), and the Program of Introducing Talents of Discipline to Universities, Plan 111 (No.B16009).

%% file: pages/sections/Appendix.tex
\appendix

\section{Dataset Construction Details}
\label{app:dataset_details}
\paragraph{Preprocessing.}
We retain source sentences containing 3--40 words and
remove exact duplicates and near duplicates with a
source-sentence Jaccard similarity of at least 0.7.
Deduplication is performed before dataset splitting.
To filter mismatched image--text pairs, the paired source
sentence must rank within the top 0.5\% of candidate texts
for its image under SigLIP similarity.

\paragraph{Automatic annotation and adjudication.}
Qwen-Max and DeepSeek-V4-Pro independently assess textual
sufficiency without access to the paired image. Their exact
three-way label agreement is 79.04\%. GPT-5.5 adjudicates
disagreements using the scoring rubric in
Table~\ref{tab:gpt_rubric}. All thresholds are set
heuristically and kept fixed throughout data construction.

The higher-scoring annotation is automatically accepted
when its total score is at least 11 and its lead is at
least 2 points. If the top annotation reaches 11 but its
lead is below 2, the instance is referred for human review.
Instances whose top annotation scores below 11 are
discarded.

\paragraph{Visual evidence verification.}
For each textually unresolved instance, SigLIP assigns
image--description compatibility scores in $[0,1]$.
An instance is automatically accepted as
\texttt{vision\_needed} when the highest score is at least
0.60 and its margin over the runner-up is at least 0.08.
Cases meeting the score threshold but not the margin
threshold undergo human review, while those below the
score threshold are discarded.

\paragraph{Human quality control.}
Two computer science master's students, both native Chinese
speakers proficient in English, independently review the
borderline cases and all 3,300 instances entering final
screening. For each instance, they inspect the source
sentence, paired image, proposed ambiguity, candidate
interpretations and translations, and supporting textual
or visual evidence. Each reviewer independently records a
pass/fail decision before discussion.

The raw pass/fail agreement is 77\%. Disagreements are
resolved through discussion. Labels are corrected when the
available evidence supports a clear revision; otherwise,
the instance is removed. Final screening removes 247
instances (7.48\%) and retains 3,053. 

\begin{table}[t]
    \centering
    \caption{Scoring rubric for GPT-based adjudication.}
    \label{tab:gpt_rubric}

    \small
    \setlength{\tabcolsep}{5pt}
    \renewcommand{\arraystretch}{1.08}

    \begin{tabular*}{\columnwidth}{
        @{\extracolsep{\fill}}lc@{}
    }
        \toprule
        Scoring dimension & Range \\
        \midrule
        Label validity               & 1--5 \\
        Ambiguity interpretation     & 1--5 \\
        Textual evidence sufficiency & 1--3 \\
        Reasoning consistency        & 1--3 \\
        \midrule
        Total                        & 4--16 \\
        \bottomrule
    \end{tabular*}
\end{table}

\section{Cross-Backend Evaluation}
\label{app:cross_backend}
\begin{table*}[t]
    \centering
    \caption{
        Overall and label-wise translation results
        with additional backends.
        Column definitions follow Table~\ref{tab:translation_results}.
        The same trained XLM-R router is reused across
        all backends without retraining.
        PreVIR uses a visual-call rate of 46.22\%,
        while Gain-Oracle is a diagnostic upper bound
        evaluated at its backend-specific native call rate.
    }
    \label{tab:additional_backends}

    \small
    \setlength{\tabcolsep}{3pt}
    \renewcommand{\arraystretch}{1.10}

    \begin{tabular*}{\textwidth}{
        @{\extracolsep{\fill}}lrrrrrrrrr@{}
    }
        \toprule
        & \multicolumn{4}{c}{BLEU$\uparrow$}
        & \multicolumn{4}{c}{COMET$\uparrow$}
        & \multicolumn{1}{c}{} \\
        \cmidrule(lr){2-5}
        \cmidrule(lr){6-9}

        Strategy
        & Overall & No amb. & Text res. & Vision req.
        & Overall & No amb. & Text res. & Vision req.
        & Vision Call \\
        \midrule

        \multicolumn{10}{l}{\textbf{InternVL3.5-8B}} \\
        \addlinespace[2pt]

        Gain-Oracle
        & 48.80 & 54.66 & 48.59 & 41.72
        & 88.31 & 90.00 & 88.73 & 87.02
        & 45.06\% \\

        Gold-label Router
        & 46.78 & -- & -- & --
        & 87.09 & -- & -- & --
        & 47.09\% \\

        Text-only
        & 44.92 & 51.16 & 47.65 & 34.28
        & 84.96 & 88.27 & 87.19 & 81.75
        & 0.00\% \\

        Always-Multimodal
        & 45.96 & 50.37 & 45.62 & 40.54
        & 87.34 & 88.92 & 87.42 & 86.29
        & 100.00\% \\

        PreVIR (Ours)
        & 46.41 & 50.99 & 48.19 & 39.01
        & 86.91 & 88.59 & 87.33 & 85.63
        & 46.22\% \\

        \midrule
        \multicolumn{10}{l}{\textbf{LLaVA-1.5-7B}} \\
        \addlinespace[2pt]

        Gain-Oracle
        & 29.60 & 32.00 & 30.00 & 25.06
        & 81.80 & 83.19 & 80.70 & 81.44
        & 43.60\% \\

        Gold-label Router
        & 27.36 & -- & -- & --
        & 79.24 & -- & -- & --
        & 47.09\% \\

        Text-only
        & 27.20 & 29.31 & 28.81 & 22.54
        & 78.32 & 80.88 & 77.87 & 76.89
        & 0.00\% \\

        Always-Multimodal
        & 26.96 & 28.16 & 27.98 & 22.89
        & 78.38 & 77.95 & 77.98 & 78.85
        & 100.00\% \\

        PreVIR (Ours)
        & 27.14 & 29.01 & 28.39 & 22.91
        & 78.75 & 80.21 & 77.56 & 78.38
        & 46.22\% \\

        \midrule
        \multicolumn{10}{l}{
            \textbf{LLaVA-OneVision-1.5-8B}
        } \\
        \addlinespace[2pt]

        Gain-Oracle
        & 47.15 & 51.32 & 49.13 & 39.88
        & 88.48 & 89.55 & 88.82 & 87.62
        & 38.95\% \\

        Gold-label Router
        & 45.21 & -- & -- & --
        & 87.25 & -- & -- & --
        & 47.09\% \\

        Text-only
        & 43.79 & 49.13 & 46.56 & 33.75
        & 85.43 & 88.18 & 87.28 & 82.78
        & 0.00\% \\

        Always-Multimodal
        & 45.21 & 48.72 & 47.45 & 38.47
        & 87.57 & 88.81 & 87.83 & 86.65
        & 100.00\% \\

        PreVIR (Ours)
        & 44.87 & 49.26 & 46.48 & 37.26
        & 87.18 & 88.59 & 87.56 & 86.09
        & 46.22\% \\

        \bottomrule
    \end{tabular*}
\end{table*}

We further evaluate PreVIR with InternVL3.5-8B, LLaVA-1.5-7B, and LLaVA-OneVision-1.5-8B on the same 344-instance test set. The trained XLM-R router is reused without backend-specific retraining, yielding identical routing decisions and a visual-call rate of 46.22\%. Table \ref{tab:additional_backends} reports overall and label-wise translation results.

Across these additional backends, PreVIR improves COMET over Text-only by 0.43–1.96 points. Compared with Always-Multimodal, COMET decreases by 0.43 points on InternVL3.5 and 0.39 points on LLaVA-OneVision, while increasing by 0.37 points on LLaVA-1.5. On the vision\_needed subset, COMET gains over Text-only range from 1.49 to 3.88 points. These results support the reuse of the learned routing policy across the evaluated backends, although the resulting translation gains remain backend-dependent.

\section{Statistical Reliability}
\label{app:reliability}
Table \ref{tab:paired_significance} reports paired score differences, 95\% confidence intervals, and two-sided p-values with Holm correction across the four comparisons. PreVIR significantly improves COMET over Text-only by 1.72 points (95\% CI: [1.01, 2.48]; adjusted p < 0.001), whereas the BLEU improvement is not statistically significant. Compared with Always-Multimodal, neither metric shows a statistically significant difference, while PreVIR uses vision for only 46.22\% of inputs. However, these results do not establish equivalence or non-inferiority. The intervals quantify test-sample uncertainty for the evaluated system outputs.

\begin{table}[t]
    \centering
    \caption{
        Paired comparisons on the 344-instance test set.
        $\Delta$ denotes PreVIR minus the baseline.
        We report pointwise 95\% confidence intervals
        for the differences and two-sided $p$-values
        adjusted using Holm's procedure across all
        four comparisons.
    }
    \label{tab:paired_significance}

    \small
    \setlength{\tabcolsep}{3pt}
    \renewcommand{\arraystretch}{1.08}

    \begin{tabular*}{\columnwidth}{
        @{\extracolsep{\fill}}lrrr@{}
    }
        \toprule
        Metric & $\Delta$ & 95\% CI & $p_{\mathrm{Holm}}$ \\
        \midrule
        \multicolumn{4}{l}{\textit{PreVIR vs.\ Text-only}} \\
        BLEU
        & $+0.97$
        & $[-0.13,\ 2.09]$
        & $0.2916$ \\
        COMET
        & $+1.72$
        & $[1.01,\ 2.48]$
        & $\mathbf{0.0004}$ \\
        \midrule
        \multicolumn{4}{l}{\textit{PreVIR vs.\ Always-MM}} \\
        BLEU
        & $-0.40$
        & $[-2.05,\ 1.25]$
        & $0.6333$ \\
        COMET
        & $-0.32$
        & $[-0.75,\ 0.08]$
        & $0.2990$ \\
        \bottomrule
    \end{tabular*}
\end{table}

\section{Practical Efficiency}
\label{app:efficiency}

\paragraph{Experimental Setup.}
We benchmark PreVIR and Always-Multimodal on the
344-instance test set, using the PreVIR routing decisions
from Table~\ref{tab:translation_results}
(46.22\% visual calls).
Both methods use Qwen3.5-9B in BF16 with vLLM 0.18.0
on a single NVIDIA A800 80GB GPU.
The router batch size is fixed at 32.
PreVIR first routes all inputs, then translates the
text-only and multimodal subsets sequentially.
We vary the translation request batch size from 2 to 48,
using the same value for both methods.
Images are resized offline to at most 307,200 pixels,
and decoding is greedy with a maximum of 128 output tokens.
Timing includes routing (for PreVIR), online image
processing, and translation generation. We use throughput to quantify the efficiency gains. The relative improvement over Always-Multimodal is $(\overline{T}_{\mathrm{Always\text{-}MM}} / \overline{T}_{\mathrm{PreVIR}} - 1) \times 100\%$.

\begin{figure}[t]
    \centering
    \includegraphics[width=\columnwidth]{
        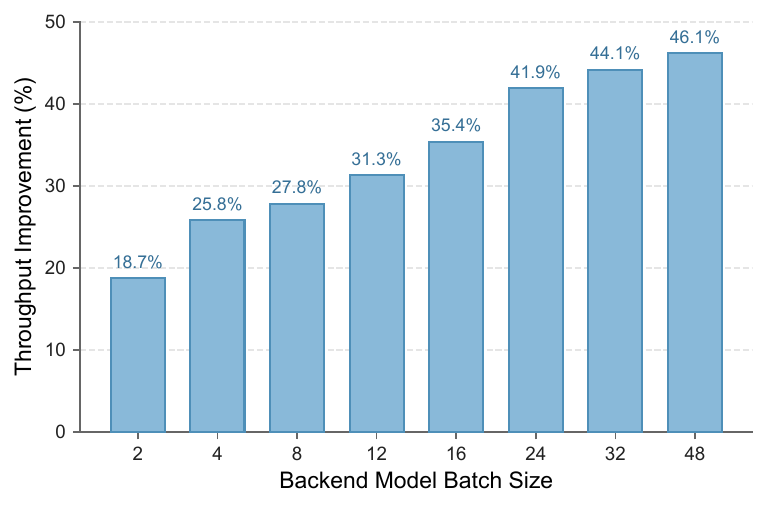
    }
    \caption{
        Throughput improvement of PreVIR over
        Always-Multimodal at matched request batch sizes,
        computed from mean total runtimes including
        router overhead.
        Runtime variability is reported in
        Table~\ref{tab:efficiency_runtime}.
    }
    \label{fig:throughput_improvement}
\end{figure}

\paragraph{Results.}
Figure~\ref{fig:throughput_improvement} shows throughput
improvements of 18.7--46.1\% across the evaluated request
batch sizes.
Table~\ref{tab:efficiency_runtime} reports the corresponding
runtimes and their variability.
Routing takes only 0.238--0.258 seconds per test-set pass,
accounting for 0.6--2.6\% of PreVIR's total runtime.
At a request batch size of 48, PreVIR reduces total runtime
from 13.592 to 9.301 seconds, a 31.6\% reduction.
Although the additional router incurs its own cost, the overall system
still delivers practical efficiency gains that far exceed this cost.

\begin{table}[t]
    \centering
    \caption{
         Processing time for the complete 344-instance test set,
     reported as mean $\pm$ standard deviation over five
    repeated runs. Always-MM and PreVIR are reported in
    seconds, whereas Router is reported in milliseconds.
    PreVIR includes both routing and translation.
    Batch denotes the translation request batch size.
    }
    \label{tab:efficiency_runtime}

    \small
    \setlength{\tabcolsep}{3pt}
    \renewcommand{\arraystretch}{1.08}

    \begin{tabular*}{\columnwidth}{
        @{\extracolsep{\fill}}crrr@{}
    }
        \toprule
        Batch & Always-MM (s) & PreVIR (s) & Router (ms) \\
        \midrule
        2
        & $48.45 \pm 0.16$
        & $40.81 \pm 0.17$
        & $255 \pm 5$ \\

        4
        & $32.88 \pm 0.06$
        & $26.14 \pm 0.02$
        & $247 \pm 7$ \\

        8
        & $23.15 \pm 0.18$
        & $18.14 \pm 0.15$
        & $253 \pm 9$ \\

        12
        & $19.76 \pm 0.04$
        & $15.05 \pm 0.05$
        & $252 \pm 6$ \\

        16
        & $17.99 \pm 0.05$
        & $13.30 \pm 0.07$
        & $238 \pm 10$ \\

        24
        & $15.815 \pm 0.132$
        & $11.147 \pm 0.081$
        & $241 \pm 16$ \\

        32
        & $14.731 \pm 0.034$
        & $10.223 \pm 0.087$
        & $258 \pm 9$ \\

        48
        & $13.592 \pm 0.068$
        & $9.301 \pm 0.042$
        & $245 \pm 17$ \\
        \bottomrule
    \end{tabular*}
\end{table}